\documentclass[a4paper]{article}

\usepackage{INTERSPEECH_v2}
\usepackage{amsmath,graphicx}
\usepackage{amssymb,amsmath,epsfig}
\usepackage{lipsum,color,multirow,url}
\usepackage{arydshln, algorithm, algpseudocode}
\usepackage{paralist}
\usepackage{enumerate}

\title{Topic Identification for Speech without ASR\thanks{This work was partially supported by DARPA LORELEI Grant N\b{o} HR0011-15-2-0024, NSF Grant N\b{o} CRI-1513128, and IARPA Contract N\b{o} 2012-12050800010.}}
\name{Chunxi Liu$^1$,  Jan Trmal$^{1,2}$, Matthew Wiesner$^1$, Craig Harman$^1$, Sanjeev Khudanpur$^{1,2}$}
\address{
  $^1$Center for Language and Speech Processing, The Johns Hopkins University, USA\\
  $^2$Human Language Technology Center of Excellence, The Johns Hopkins University, USA}
\email{\{chunxi, yenda, wiesner, khudanpur\}@jhu.edu, craig@craigharman.net}

\begin{document}
\maketitle
\begin{abstract}
Modern topic identification (topic ID) systems for speech use automatic speech recognition (ASR) to produce speech transcripts, and perform supervised classification on such ASR outputs. However, under resource-limited conditions, the manually transcribed speech required to develop standard ASR systems can be severely limited or unavailable. In this paper, we investigate alternative unsupervised solutions to obtaining tokenizations of speech in terms of a vocabulary of automatically discovered word-like or phoneme-like units, without depending on the supervised training of ASR systems. Moreover, using automatic phoneme-like tokenizations, we demonstrate that a convolutional neural network based framework for learning spoken document representations provides competitive performance compared to a standard bag-of-words representation, as evidenced by comprehensive topic ID evaluations on both single-label and multi-label classification tasks.
\end{abstract}
\section{Introduction}


Topic identification (topic ID) on speech aims to identify the topic(s) for given speech recordings, referred to as spoken documents, where the topics are a predefined set of classes or labels. This task is typically formulated as a three-step process.
First, speech is tokenized into words or phones by automatic speech recognition (ASR) systems~\cite{Hazen:2011:MCE_topic_ID}, or by limited-vocabulary keyword spotting~\cite{wintrode2014limited}. 
Second, standard text-based processing techniques are applied to the resulting tokenizations, and produce a vector representation for each spoken document, typically a bag-of-words multinomial representation, or a more compact vector given by probabilistic topic models~\cite{blei2003latent, may2015topic}. 
Finally, topic ID is performed on the spoken document representations by supervised training of classifiers, such as Bayesian classifiers and support vector machines (SVMs). 


However, in the first step, training the ASR system required for tokenization itself requires transcribed speech and pronunciations. In this paper, we focus on a difficult and realistic scenario where the speech corpus of a test language
is annotated only with a minimal number of topic labels, i.e., no manual transcriptions or dictionaries for building an ASR system are available.
We aim to exploit approaches that enable topic ID on speech without any knowledge of that language other than the topic annotations. 

In this scenario, while previous work demonstrates that the cross-lingual phoneme recognizers can produce reasonable speech tokenizations~\cite{siu2014unsupervised, kesiraju2017empirical}, the performance is highly dependent on the language and environmental condition (channel, noise, etc.) mismatch between the training and test data. 
Therefore, we focus on unsupervised approaches that operate directly on the speech of interest. Raw acoustic feature-based unsupervised term discovery (UTD) is one such approach that aims to identify and cluster repeating word-like units across speech based around segmental dynamic time warping (DTW)~\cite{park2008unsupervised, jansen2011efficient}.~\cite{dredze2010nlp} shows that using the word-like units from UTD for spoken document classification can work well; however, 
the results in~\cite{dredze2010nlp} are limited since
the acoustic features on which UTD is performed are produced by acoustic models trained from the transcribed speech of its evaluation corpus. In this paper, we investigate UTD-based topic ID performance when UTD operates on language-independent speech representations extracted from multilingual bottleneck networks trained on languages other than the test language~\cite{liu2017empirical}.
Another alternative to producing speech tokenizations without language dependency is the model-based approach, i.e., unsupervised learning of hidden Markov model (HMM) based phoneme-like units from untranscribed speech. We exploit the Variational Bayesian inference based acoustic unit discovery (AUD) framework in~\cite{ondel2016variational} that allows parallelized large-scale training. In topic ID tasks, such AUD-based systems have been shown to outperform other systems based on cross-lingual phoneme recognizers~\cite{kesiraju2017empirical}, and this paper aims to further investigate how the performance compares among UTD, AUD and ASR based systems. 


Moreover, after the speech is tokenized, these works~\cite{Hazen:2011:MCE_topic_ID, wintrode2014limited, siu2014unsupervised, kesiraju2017empirical, dredze2010nlp, liu2017empirical} are limited to using bag-of-words features as spoken document representations. 
While UTD only identifies relatively long (0.5 -- 1 sec) repeated terms, AUD/ASR enables full-coverage segmentation of continuous speech into 
a sequence of units/words, and such a resulting temporal sequence enables another feature learning architecture based on convolutional neural networks (CNNs)~\cite{collobert:08}; instead of treating the sequential tokens as a bag of acoustic units or words, the whole token sequence is encoded as concatenated continuous vectors, and followed by convolution and temporal pooling operations that capture the local and global dependencies. Such continuous space feature extraction frameworks have been used in various language processing tasks like spoken language understanding~\cite{Xu:13, liu2015deep} and text classification~\cite{kim2014convolutional, zhang2015character}. 
However, three questions are worth investigating in our AUD-based setting: (i) if such a CNN-based framework can perform as well on noisy automatically discovered phoneme-like units as on orthographic words/characters, (ii) if pre-trained vectors of phoneme-like units from \emph{word2vec}~\cite{mikolov2013efficient} provide superior performance to random initialization as evidenced by the word-based tasks, and (iii) if CNNs are still competitive in low-resource settings of hundreds to two-thousand training exemplars, rather than the large/medium sized datasets as in previous work~\cite{kim2014convolutional, zhang2015character}.

Finally, incorporating the different tokenization and feature representation approaches noted above, we perform comprehensive topic ID evaluations on both single-label and multi-label spoken document classification tasks. 


\vspace{-0.1cm}
\section{Unsupervised tokenizations of speech}
\vspace{-0.1cm}

\subsection{Unsupervised term discovery (UTD)}
\label{subsec:UTD}
UTD aims to automatically identify and cluster repeated terms (e.g. words or phrases) from speech. To circumvent the exhaustive DTW-based search limited by $\mathcal{O}(n^2)$ time~\cite{park2008unsupervised}, we exploit the scalable UTD framework in the Zero Resource Toolkit (ZRTools)~\cite{jansen2011efficient}, which permits search in $\mathcal{O}(n \log n)$ time. We briefly describe the UTD procedures in ZRTools by four steps below, and full details can be found in~\cite{jansen2011efficient}. 
\vspace{-0.2cm}
\begin{enumerate}
\item Construct the sparse approximate acoustic similarity matrices between pairs of speech utterances. 
\item Identify word repetitions via fast diagonal line search and segmental DTW.
\item The resulting matches are used to construct an acoustic similarity graph, where nodes represent the matching acoustic segments and edges reflect DTW distances. 
\item Threshold the graph edges, and each connected component of the graph is a cluster of acoustic segments, which produces a corresponding term (word/phrase) category.  
\end{enumerate}
\vspace{-0.2cm}
Finally, the cluster of each discovered term category consists of a list of term occurrences. 

Note that in the third step above, the weight on each graph edge can be exact DTW-based similarity, or other similarity based on heuristics more than DTW distance. 
For example, 
we investigate an implementation in ZRTools, where
a separate logistic regression model is used to rescore the similarity between identified matches by determining how likely the matching pair is the same underlying word/phrase and is not a filled pause (e.g. ``um-hum'' and ``yeah uh-huh'' in English). Filled pauses tend to be acoustically stationary with more phone repeats and thus would match throughout the acoustic similarity matrix, whereas a contentful word (without too many phone repeats) tend to concentrate around the main diagonal; thus, the features in logistic regression contain the numbers of matrix elements in diagonal bands in progressive steps away from the main diagonal. Feature weights are learned using a portion of transcribed speech with reference transcripts, and the resulting model can be used for language-independent rescoring. 

\vspace{-0.1cm}
\subsection{Acoustic unit discovery (AUD)}



We exploit the nonparametric Bayesian AUD framework in~\cite{ondel2016variational} based on variational inference, rather than the maximum likelihood training in~\cite{siu2014unsupervised} which may oversimplify the parameter estimations, nor the Gibbs Sampling training in~\cite{lee2012nonparametric} which is not amenable to large scale applications.
Specifically, a phone-loop model is formulated 
where each phoneme-like unit is modeled as an HMM with a Gaussian mixture model of output densities (GMM-HMM). Under the Dirichlet process framework, we consider the phone loop as an infinite mixture of GMM-HMMs, and the mixture weights are based on the stick-breaking construction of Dirichlet process. The infinite number of units in the mixture is truncated in practice,  giving zero mixture weight to any unit beyond some large count. 
We treat such mixture of GMM-HMMs as a single unified HMM 
and thus the segmentation of the data is performed using standard forward-backward algorithm. Training is fully unsupervised and parallelized; after a fixed number of training iterations, we use Viterbi decoding algorithm to obtain acoustic unit tokenizations of the data.

\begin{figure}[t]
  \centering
  \includegraphics[width=6.6cm, scale=0.80]{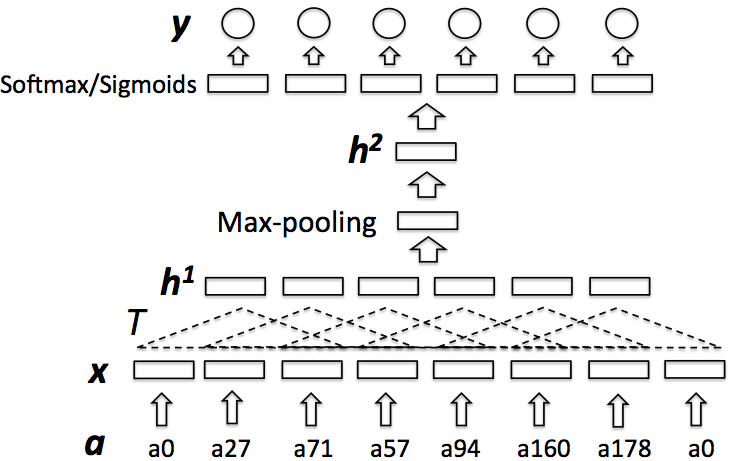}
  \vspace{-0.32cm}
  \caption{CNN-based framework that operates on automatically discovered acoustic units.}
  \label{fig:cnn}
\vspace{-0.68cm}
\end{figure}
\section{Learning document representations}

\subsection{Bag-of-words representation}

After we obtain the tokenizations of speech by either UTD or AUD, each spoken document is represented 
by a vector of unigram occurrence counts over discovered terms, or a vector of $n$-gram counts over acoustic units, respectively. 
Each feature vector can be further scaled by inverse document frequency (IDF), producing a TF-IDF feature.

\subsection{Convolutional neural network-based representation}
\label{subsec:cnn}

AUD enables full-coverage tokenization of continuous speech into a sequence of acoustic units, which we can exploit in a CNN-based framework to learn a vector representation for each spoken document. As shown in Figure~\ref{fig:cnn}, in an acoustic unit sequence \textbf{a} of length $m$, each unit $a_i $, $1 \leq i \leq m$, is encoded as a fixed dimensional continuous vector, and the whole sequence \textbf{a} is represented as a concatenated vector \textbf{x}. 
A shared convolutional feature transform $T$ spans a fixed-sized $n$-gram window, $n\ll m$, and slides over the whole sequence. Then the hidden feature layer $ \textbf{h}^1$ with nonlinearities consists of each feature vector $h_{i}^1$ extracted from the shared convolutional window centered at each acoustic unit position $i$. 
Max-pooling is performed on top of each $h_{i}^1$, $1 \leq i \leq m$, to obtain a fixed-dimensional vector representation for the whole sequence \textbf{a}, i.e., a vector representation of the whole spoken document, followed by another hidden layer  $\textbf{h}^2$  and a final output layer. Note that this framework needs supervision for training; e.g., the output layer can be a softmax function for single-label classification, and the whole model is trained with categorical cross-entropy loss. 

Also, 
the vector representation of each unique acoustic unit can be randomly initialized, or pre-trained from other tasks. Specifically, we apply the \emph{skip-gram} model of word2vec~\cite{mikolov2013distributed} to pre-train one embedding vector for each acoustic unit, based on the hierarchical softmax with Huffman codes.

\section{Supervised document classification}
\label{sec:classification}

\subsection{Single-label classification}
\label{subsec:single_label}

For the bag-of-words representation, 
we use a stochastic gradient descent (SGD) based linear SVM~\cite{shalev2007pegasos, scikit-learn} with hinge loss and $\mathcal{L}^1$/$\mathcal{L}^2$ norm regularization. For the CNN-based framework, we use a softmax function in the output layer for classification as described in Section~\ref{subsec:cnn}.

\subsection{Multi-label classification}
\label{subsec:multi_label}

In the setting where each spoken document can be associated with multiple topics/labels, we proceed to perform a multi-label classification task. The baseline approach is the binary relevance method, which independently trains one binary classifier for each label, and the spoken document is evaluated by each classifier to determine if the respective label applies to it. Specifically, we use a set of SVMs (Section~\ref{subsec:single_label}), one for each label, on the bag-of-words features.

To adapt the CNN-based framework for multi-label classification, we replace the softmax in the output layer with a set of sigmoid output nodes, one for each label, as shown in Figure~\ref{fig:cnn}. Since a sigmoid naturally provides output values between $0$ and $1$, we train the neural network (NN) to minimize the binary cross entropy loss defined as
$ l(\Theta, (\textbf{x}, \textbf{y})) = -  \sum_{k = 1}^{K} ( y_k \log o_k +  (1 - y_k)  \log (1 - o_k)   ) $,
where $\Theta$ denotes the NN parameters, \textbf{x} is the feature vector of acoustic unit sequence, \textbf{y} is the target vector of labels, $o_k$ and $y_k$ are the output and the target for label $k$, and the number of unique labels is $K$.

\section{Experiments}

\subsection{Single-label classification}

\subsubsection{Experimental setup}


For our single-label topic classification experiments, we use the Switchboard Telephone Speech Corpus~\cite{godfrey2010switchboard}, a collection of two-sided telephone conversations.
We use the same development (dev) and evaluation (eval) data sets as in~\cite{dredze2010nlp, liu2017empirical}. 
Each whole conversation has two sides and one single topic, and topic ID is performed on each individual-side speech (i.e., each side is seen as one single spoken document). 
In the 35.7 hour dev data, there are 360 conversation sides evenly distributed across six different topics (recycling, capital punishment, drug testing, family finance, job benefits, car buying), i.e., each topic has equal number of 60 sides. In the 61.6 hour eval data, there are another different six topics (family life, news media, public education, exercise/fitness, pets, taxes) evenly distributed across 600 conversation sides.
Algorithm design choices are explored through experiments on dev data.
We use manual segmentations provided by the Switchboard corpus to produce utterances with speech activity, and UTD and AUD are operating only on those utterances.  

For UTD, we use the ZRTools~\cite{jansen2011efficient} implementation with the default parameters except that, we use cosine similarity threshold $\delta = 0.5$, and vary the diagonal median filter duration $\kappa$ over $\{0.6, 0.7\}$; we try both the exact DTW-based similarity and the rescored similarity as described in Section~\ref{subsec:UTD}, and tune the similarity threshold (used to partition the graph edges) over $\{0.85, 0.88, 0.90, 0.92\}$.
For AUD, the unsupervised training is performed only on the dev data (10 iterations); after training, we use the learned models to decode both dev and eval data set, and obtain the acoustic unit tokenizations. 
We use truncation level 200, which implies maximum 200 different acoustic units can be learned from the corpus. For each acoustic unit, we use a 3-state HMM with 2 Gaussians per state. For the stick-breaking construction of Dirichlet process,  we vary the concentration parameter $\gamma$ over $\{1.0, 10.0\}$, and other hyperparameters are the same as~\cite{ondel2016variational}.    

The acoustic features on which UTD and AUD operate are extracted using the same multilingual bottleneck (BN) network as described in~\cite{liu2017empirical} with Kaldi toolkit~\cite{povey2011kaldi}.  
We conduct the multilingual BN training with 10 language collections (Assamese, Bengali, Cantonese, Haitian, Lao, Pashto, Tamil, Tagalog, Vietnamese and Zulu) -- 10 hours of transcribed speech per language. Complete specifications can be found in~\cite{liu2017empirical}. 


For SVM-based classification, we use the bag of discovered term unigrams, or bag of acoustic unit trigrams. 
On dev data, we try using the features of raw counts or scaled by IDF, SVM regularization tuned over $\mathcal{L}^1$/$\mathcal{L}^2$ norm, regularization constant tuned over $\{0.001, 0.0001\}$, and SGD epochs tuned over $\{30, 50\}$. We further normalize each feature to $\mathcal{L}^2$ norm unit length. 
Each experiment is a run of 10-fold cross validation (CV) on the 360 conversation sides of dev data, or on the 600 sides of eval data, respectively. 
Note that our data size here is relatively small (only 360 or 600) and the SGD training may give high variance in the performance~\cite{zhang2015sensitivity}. Therefore, to report classification accuracy for each configuration (when varying features or models), we repeat each CV experiment 5 times, where each experiment again is a run of 10-fold CV; then for each configuration, the mean and standard deviation of 5 experiments is reported. 

For CNN-based classification, we use the same strategy to report classification accuracy, i.e., repeating experiments 5 times (where each time is a 10-fold CV) for each CNN configuration. Note that the respective 10 folds of both dev and eval data sets are fixed the same for all the SVM and CNN experiments. Additionally, for each 10-fold CV experiment, instead of training on 9 folds and testing on the remaining 1 fold as in SVM, we use 8 folds for CNN training, leave another 1 fold out as validation data; after training each CNN model for up to 100 epochs, the model with the best accuracy on the validation data is used for evaluation on the test set.
The acoustic unit sequence (as CNN inputs) are zero-padded to the longest length in each dataset.  
We implemented the CNNs in Keras~\cite{chollet2015keras} with Theano~\cite{2016arXiv160502688short} backend.
CNN architectures are determined through experiments on dev data. 
For SGD training we use the Adadelta optimizer~\cite{zeiler2012adadelta} and mini-batch size 18. 
The $n$-gram window size of each convolutional feature transform $T$ is 7. The size of each hidden feature vector $h^1_i$ (extracted from the transform $T$) is 1024, with rectified linear unit (ReLU) nonlinearities. 
Thus, after max-pooling over time, we have a 1024-dimensional vector again, which then goes through another hidden layer $\textbf{h}^2$ (also set as 1024-dimensional with ReLU) and finally into a softmax. Dropout~\cite{hinton2012} rate 0.2 is used at each layer. 

When we initialize the vector representation of each acoustic unit with a set of pre-trained vectors (instead of random initializations), we apply the skip-gram model of word2vec~\cite{mikolov2013distributed} to the acoustic unit tokenizations of each data set. We use the \emph{gensim} implementation~\cite{rehurek_lrec}, which includes a vector space of embedding dimension 50 (tuned over $\{50, 80\}$), a skip-gram window of size 5, and SGD over 20 epochs. 


\subsubsection{Results on Switchboard}

Table~\ref{tab:result1} shows the topic ID results on Switchboard. For UTD-based classifications, we find that the default rescoring in ZRTools~\cite{jansen2011efficient} which is designed to filter out the filled pauses produces comparable performance to the raw DTW similarity scores, but the rescoring can result in much faster connected-component clustering (Section~\ref{subsec:UTD}).
Note that this rescoring model is estimated using a portion of transcribed Switchboard, 
but it is still a legitimate language-independent UTD approach while operating on languages other than English.
While a diagonal median filter duration $\kappa$ of $0.6$ or $0.7$ gives similar results,
$\kappa = 0.7$ produces longer but fewer terms, giving more sparse feature representations. Therefore, we proceed with rescoring and $\kappa = 0.7$ in the following UTD experiments (Section~\ref{subsec:multi_label_exp}).

\setlength{\tabcolsep}{0.12cm}
\begin{table}[t]
\vspace{-0.01cm}
\caption{\label{tab:result1} {\it Single-label topic ID accuracies on Switchboard.}}
\vspace{-1mm}
\centerline{ 
\begin{tabular}{c|c|c|c}
\hline \hline
Dataset           & Feature   &  Model  & Accuracy       \\
\hline \hline
\multirow{7}{*}{Dev}     &   UTD        &  SVM   &     0.863    $\pm$  0.010    \\
                                    &   UTD w/ rescoring          & SVM   &     0.876     $\pm$  0.008   \\         \cline{2-4}     
                           &                                      &  SVM                         &  0.682 $\pm$  0.007  \\ 
                           &   AUD, \# units 184       & CNN                          &  0.657  $\pm$  0.017  \\ 
                           &                                      &   CNN w/ word2vec   &  0.728 $\pm$  0.011   \\   \cdashline{2-4}[2.0pt/0.5pt]
                           &                                      & SVM                          &  0.686 $\pm$  0.005  \\ 
                           &   AUD,  \# units 199      & CNN                          & 0.749  $\pm$  0.008   \\ 
                           &                                      & CNN w/ word2vec     & \textbf{0.763}  $\pm$  0.011  \\  \cdashline{2-4}[2.0pt/0.5pt] 
\hline \hline
\multirow{7}{*}{Eval}    &  UTD          &  SVM  &   0.851   $\pm$  0.003    \\
                                    &  UTD    w/ rescoring         &  SVM  &   0.875    $\pm$  0.003    \\        \cline{2-4}  
                           &                                      &   SVM                      &  0.710     $\pm$  0.005   \\ 
                           &   AUD, \# units = 184    &   CNN                      &  0.708    $\pm$  0.013    \\ 
                           &                                      &   CNN w/ word2vec &  0.762   $\pm$  0.007   \\   \cdashline{2-4}[2.0pt/0.5pt]
                           &                                      &   SVM                      &     0.700         $\pm$  0.005   \\ 
                           &   AUD, \# units = 199    &   CNN                      &  0.690    $\pm$  0.015   \\ 
                           &                                      &   CNN w/ word2vec  & \textbf{0.767} $\pm$  0.013   \\  \cdashline{2-4}[2.0pt/0.5pt] 
\hline \hline
\end{tabular}}
\vspace{-0.6cm}
\end{table}

For AUD-based classifications, CNN without word2vec pre-training usually gives comparable results with SVM; however, using word2vec pre-training, CNN substantially outperforms the competing SVM in all cases. Also as the concentration parameter $\gamma$ in AUD increases from $1.0$ to $10.0$ (yielding less concentrated distributions), we have more unique acoustic units in the tokenizations of both data sets, from 184 to 199, and $\gamma = 10.0$ usually produces better results than $\gamma = 1.0$. 

\subsection{Multi-label classification}
\label{subsec:multi_label_exp}

\subsubsection{Experimental setup}


We further evaluate our topic ID performance on the speech corpora of three languages released by the DARPA LORELEI (Low Resource Languages for Emergent Incidents) Program. 
For each language there are a number of audio speech files, and each speech file is cut into segments of various lengths (up to 120 seconds).
Each speech segment is seen as either in-domain or out-of-domain. In-domain data is defined as any speech segment relating to an incident or incidents, and in-domain data will fall into a set of domain-specific categories; these categories are known as situation types, or in-domain topics. 
There are 11 situation types: ``Civil Unrest or Wide-spread Crime'', ``Elections and Politics'', ``Evacuation'', ``Food Supply'', ``Urgent Rescue'', ``Utilities, Energy, or Sanitation'', ``Infrastructure'', ``Medical Assistance'', ``Shelter'', ``Terrorism or other Extreme Violence'', and ``Water Supply''. We consider ``Out-of-domain'' as the 12th topic label, so each speech segment either corresponds to one or multiple in-domain topics, or is ``Out-of-domain''. 
We use the average precision (AP, equal to the area under the precision-recall curve) as the evaluation metric, and report both the AP across the overall 12 labels, and the AP across 11 situation types, as shown in Table~\ref{tab:result2}. For each configuration, only a single 10-fold CV result is reported, since we observe less variance in results here than in Switchboard. 
We have 16.5 hours in-domain data and 8.5 hours out-of-domain data for Turkish, 2.9 and 13.2 hours for Uzbek, and 7.7 and 7.2 hours for Mandarin. 
We use the same CNN architecture as on Switchboard but make the changes as described in Section~\ref{subsec:multi_label}. Also we use mini-batch size 30 and fix the training epochs as 100. All CNNs use word2vec pre-training. 
Additionally, we also implement another two separate topic ID baselines using the decoded word outputs from two supervised ASR systems, trained from 80 hours transcribed Babel Turkish speech~\cite{trmal2014keyword} and about 170 hours transcribed HKUST Mandarin telephone speech (LDC2005T32 and LDC2005S15), respectively.

\setlength{\tabcolsep}{0.20cm}
\begin{table}[t]
\vspace{-0.1cm}
\caption{\label{tab:result2} {\it Multi-label topic ID average precision on LORELEI languages, with the number of speech segments in parentheses.}}
\vspace{-3mm}
\centerline{ 
\begin{tabular}{c|c|c|c|c}
\hline \hline
Dataset        &      Feature       &       Model      & Overall    & In-domain topics   \\  
\hline \hline
\multirow{3}{*}{Turkish} &      UTD &   SVM   &  0.583      &    0.531   \\  
                      &     AUD     &    SVM    &   0.627  &  0.556 \\        
           (2095)       &       AUD       &       CNN    &  0.641    &    0.564   \\     
           &  ASR &      SVM     &   0.625       &   0.580     \\       
\hline \hline
\multirow{3}{*}{Uzbek} &    UTD     &     SVM    &    0.803      &    0.254     \\  
                                    &    AUD     &     SVM    &   0.791       &    0.203     \\  
     (1416)                     &    AUD     &     CNN    &   0.807       &    0.207     \\ 
\hline \hline
\multirow{3}{*}{Mandarin} &   UTD     &     SVM    &   0.444    &   0.234       \\  
                           &    AUD     &     SVM    &    0.436   &    0.220    \\     
        (724)    &  AUD     &     CNN    &  0.420    &     0.183  \\ 
         &   ASR     &     SVM     &  0.461       &      0.261    \\  
\hline \hline
\end{tabular}}
\vspace{-0.4cm}
\end{table}
\begin{figure}[t]
  \centering
  \includegraphics[height=4.66cm, width=8.0cm]{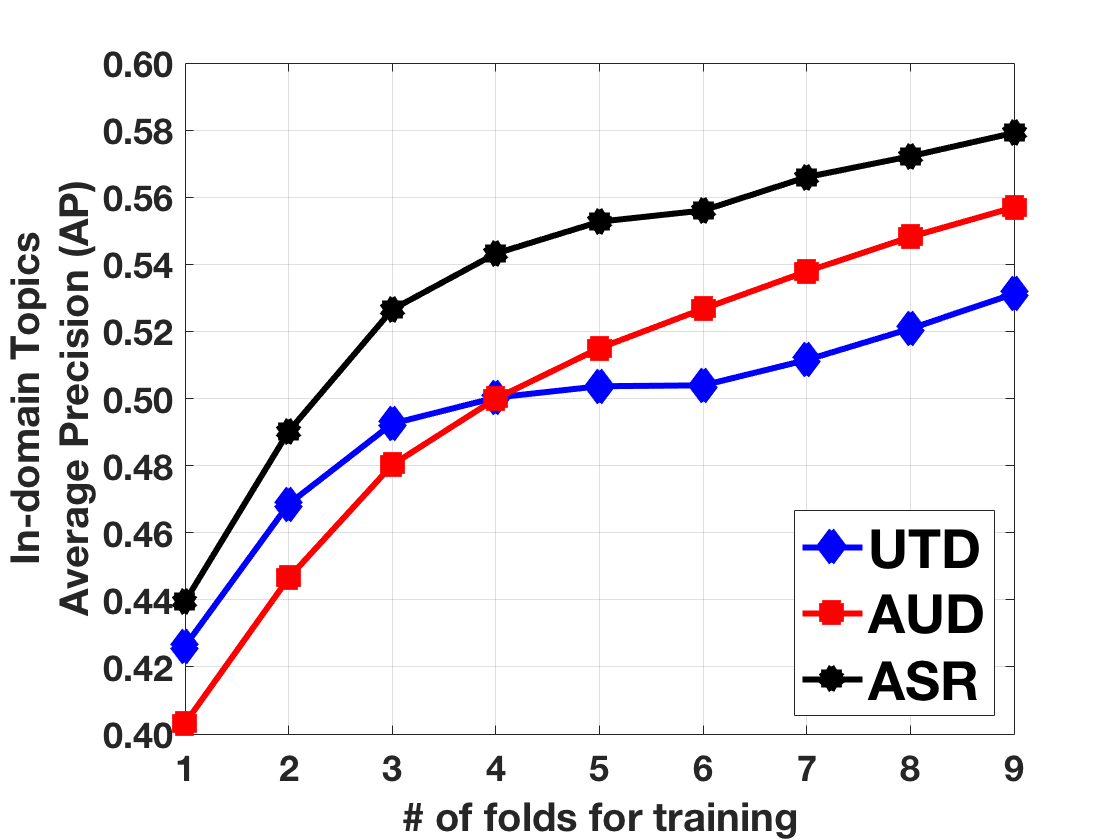}
  \vspace{-0.7cm}
  \caption{Average precision of in-domain situation types on Turkish when varying the number of folds used for training.}
  \label{fig:turkish}
\vspace{-0.7cm}
\end{figure}

\subsubsection{Results on LORELEI datasets}
\vspace{-0.1cm}

As shown in Table~\ref{tab:result2}, UTD-based SVMs are more competitive than AUD-based SVMs on the smaller corpora, i.e., Uzbek and Mandarin, while being less competitive on the larger  corpus, Turkish. 
We further investigate this behavior on each individual language by varying the amount of training data; we split the data into 10 folds, and perform 10-fold CV 9 times, varying the number of folds for training from 1 to 9. 
As illustrated in Figure~\ref{fig:turkish} for Turkish, as we use more folds for training, AUD-based system starts to be more competitive than UTD. 
Supervised ASR-based systems still give the best results in various cases, while UTD and AUD based systems give comparable performance. 

Note that CNN-based systems outperform SVMs on Turkish and Uzbek while losing on the smaller sized Mandarin, indicating more topic-labeled data is needed to enable competitive CNNs. This also indicates why CNNs on LORELEI corpora do not produce as large a gain over SVMs as on the larger sized Switchboard, since each 15-25 hour LORELEI corpus with 12 topic labels is a relatively small amount of data compared to the 35.7/61.6 hour Switchboard corpus with 6 labels. 

\section{Concluding remarks}


We have demonstrated that both UTD and AUD are viable technologies for producing effective tokenizations of speech that enable topic ID performance comparable to using standard ASR systems, while effectively removing the dependency on transcribed speech required by the ASR alternative.  
We find that when training data is severely limited the UTD-based classification is superior to AUD-based classification. As the amount of training data increases, performance improves across the board. Finally, with sufficient training data AUD-based CNNs with word2vec pre-training outperform AUD-based SVMs.



\vfill\pagebreak
\bibliographystyle{IEEEtran}

\bibliography{mybib}

\end{document}